# Design and Evaluation of Torque Compensation Controllers for a Lower Extremity Exoskeleton

Xianlian Zhou[1] and Xinyu Chen[2]
[1]New Jersey Institute of Technology, Newark, NJ 07102, alexzhou@njit.edu
[2]CFD Research Corporation, 2. Huntsville, AL 35806

## ABSTRACT

*In this paper, we present an integrated human-in-the-loop simulation paradigm for the design and evaluation of a lower extremity exoskeleton that is elastically strapped onto human lower limbs. The exoskeleton has 3 rotational DOFs on each side and weighs 23kg. Two torque compensation controllers of the exoskeleton are introduced, aiming to minimize interference and maximize assistance to human motions, respectively. Their effects on the wearer's biomechanical loadings are studied with a running motion and predicted ground reaction forces. It is found that the added weight of the passive exoskeleton substantially increases the wearer's musculoskeletal loadings. The maximizing assistance controller reduces the knee joint torque by almost a half when compared to the passive exoskeleton and the resultant torque is only 72% of that from the normal running without exoskeleton. When compared to the normal running, this controller also reduces the hip flexion and extension torques by 31% and 38%, respectively. As a result, the peak activations of the biceps short head, gluteus maximus, and rectus femoris muscles are reduced by more than a half. Nonetheless, the axial knee joint reaction force increases for all exoskeleton cases due to the added weight and higher GRFs. In summary, the results provide sound evidence of the efficacy of these two controllers on reducing the wearer's musculoskeletal loadings when*

---

[1] Corresponding author

*compared to the passive exoskeleton. And it is shown the human-in-the-loop simulation paradigm presented here can be used for virtual design and evaluation of powered exoskeletons and pave the way for building optimized exoskeleton prototypes for experimental evaluation.*

1. Introduction

To augment human performance or assist with disabilities, many exoskeletons or exosuit systems have been developed in recent years. Early efforts can date back to the 1960s' General Electric HARDIMAN project which attempted to develop the first practical powered exoskeleton for military applications. More recently, a DARPA sponsored program, called Exoskeletons for Human Performance Augmentation (EHPA), funded two promising exoskeleton systems: the Berkeley Bionics/Lockheed Martin HULC (Human Universal Load Carrier) system and the SARCOS Labs/Raytheon XOS system. The HULC system has been further developed and tested by Lockheed Martin and others to help soldiers in combat to carry a load of up to 200 pounds at a top speed of 10 miles per hour for an extended period of time. Outside the US, similar efforts have also been conducted and resulted in several exoskeleton systems for civilian or military applications such as the French RB3D's HERCULES system and the Japan Cyberdyne's HAL system [1]. Other than robotic exoskeleton systems, soft exosuit [2] also exist aiming to provide gait assistance and reduce metabolic cost of human locomotion. Compared to robotic exoskeletons, exosuit are in general lightweight but provide limited power assistance.

Because of their bulky sizes, rigid interfaces, and electric power requirements, most existing robotic exoskeleton systems face some technical challenges for their applications. Typically, a wearable exoskeleton is designed to have salient features such as: 1) fitting to individual's body shape and size, 2) lowering the metabolic cost of human locomotion, 3) preventing musculoskeletal injury or trauma during dynamic events, and 4) reducing interference and providing assistance to human motion with effective control and actuation. A recent US army experimental study conducted by Gregorczyk et al. [3] evaluated a prototype exoskeleton system similar to HULC and found the system altered the wearers' gait and increased their oxygen consumption (VO2) significantly. Therefore, a well-designed control and actuation scheme is particularly important to achieve the aforementioned features. However, this is a challenging task due to the complexity and variability of human locomotion. There are various control and actuation approaches that have been reviewed in the literature [4, 5]. For example, the HAL system [6] powers the hip and knee joints via a DC motor with harmonic drives placed directly on the joints. It utilizes skin-surface EMG and a walking pattern based control system to determine user intent in order to operate the suit. However, it was reported to take two months to optimally calibrate the system for a specific user [7]. The BLEEX (Berkeley Lower Extremity Exoskeleton) system developed by Kazerooni et al. [8, 9] features bidirectional linear hydraulic actuations for hip flexion/extension and abduction/adduction, knee flexion/extension, and ankle flexion/extension. Its controller utilizes mainly sensory information from the exoskeleton and enables the exoskeleton to balance on its own while the wearer provides a forward interaction force to guide the system during walking. Meijneke et al. [10] presented the Achilles exoskeleton, an autonomous ankle exoskeleton with high power density achieved by designing a series of elastic actuators, which consists of an electric motor and a ball-screw gear, with a carbon fiber reinforced leaf-spring as the lever-arm.

Due to the challenges in the design, control, and actuation of exoskeleton systems, it is highly desired to virtually test a prototype system before physically assembling the system in order to save material cost and labor. This requires a human-in-the-loop modeling method to simulate exoskeleton-wearer interactions, as demonstrated in several recent studies [11-14]. To study the effects of exoskeleton on the wearer's biomechanical loadings, musculoskeletal modeling software such as OpenSim [15] and AnyBody [16] have been used. For example, Zhou et al. [17] used AnyBody for the design and optimization of a spring-loaded cable-driven robotic exoskeleton. Koller et al. [18] used OpenSim to study adaptive gain proportional myoelectric controllers for a robotic ankle exoskeleton. Recently, Delp's group at Stanford University published two studies on simulating ideal (massless) assistive devices to reduce the metabolic cost of walking (with heavy loads [19]) and running [20]. In both studies, OpenSim was used to generate muscle-driven simulations of multiple subjects walking or running with massless assistive devices, which applied ideal net joint moments directly to the human joints without considering physical interaction forces between the devices and the subjects. In addition, the kinematics and the ground reaction forces (GRFs) obtained experimentally during unassisted running remained unchanged when assistances were added in the simulations.

In this paper, we present an integrated human-in-the-loop simulation paradigm for the design and evaluation of two virtual control schemes of a lower extremity exoskeleton and study their effects on the wearer's biomechanical loadings. At first, the design of a lower extremity exoskeleton is introduced with detailed description of joints and mechanism, mass and inertia, actuations, and human-device interactions. It is followed by a concise description of the overall simulation method, in which the GRFs are predicted and muscle force coordination is optimized. Then two active torque compensation controllers for the exoskeleton are proposed, aiming to 1) minimize the interaction forces between the lower limbs of the exoskeleton and the wearer and 2) provide assistive torques to maximize the help to the wearer's motion. These two controllers are evaluated with running simulations and their results on exoskeleton-wearer interaction forces and biomechanical loadings are compared with those from the normal running and the passive exoskeleton. Lastly, discussion and conclusions are presented.

## 2. METHODS

### 2.1 A LOWER EXTREMITY EXOSKELETON DESIGN

The lower extremity exoskeleton design is shown in Figure 1. It has 7 mechanical parts linked by joints: the load support frame, the exo-pelvis (L/R), the exo-femur (L/R), and the exo-tibia (L/R) (L: left; R: right). The total weight of this exoskeleton is 23kg and the mass and inertia properties of its components are listed in Table 1. Note the 3D geometrical models of these components are simple representation of the real parts and do not necessarily include all accessories such as motors, batteries, and wires. On each side, there are 3 joints: one links the load support with the exo-pelvis; one links the exo-pelvis with exo-femur, and the third one links the exo-femur with the exo-tibia. Each joint has a single rotational degree of freedom (DOF) and its joint axes are displayed in the figure. The exoskeleton also contains 6 idealized actuators (displayed as yellow cylinders) that can generate both positive and negative (or push and pull) forces, with 3 on each side. These idealized actuators can represent typical hydraulic actuators or electric motor actuators as used by Meijneke et al. [10]. They are attached at locations that allow effective actuation of each joint along their DOF. Based on their functionalities, from top to bottom, these actuators are called exo-pelvis, exo-hip, and exo-knee, respectively. Each actuator can generate an active force up to $\pm 4000N$ (positive: pull; negative: push), reachable by many hydraulic or electric motor actuators.

In Figure 2, the exoskeleton is assembled onto a whole body musculoskeletal (MSK) model with elastic straps. In what follows, a few assumptions have been made in modeling and simulation. First, the load support is assumed to be tied to the human body such that there is no relative movement. The femur straps are assumed to have different directional force responses to relative XYZ movements between the exo-femur and human femur (X: fore-aft, Y: vertical, Z: lateral). So are the tibia straps. It enables strong resistance against relative movement along the fore-aft (forward) direction and much weaker resistance along the vertical (sliding) and lateral (abduction/adduction) directions. This setup can simulate straps made of partial hard plastic or metal (e.g. at the front and back sides of the leg) and partial elastic fabric (e.g. at the inner side of the leg). It is also important considering there exist a few inconsistencies in the human hip and knee joints and their exoskeleton counterparts. The human hip joint has three rotational DOFs (flexion, abduction, and axial rotation) with the same center of rotation (COR). The exoskeleton has only two rotational DOFs for the flexion and abduction motions and their CORs are located at two distinct locations (exo-pelvis and exo-hip joints). Weaker resistance along the lateral direction will cause less force interference for abduction/adduction despite these discrepancies. The human knee joint constrains the tibia to rotate around the end of femur with its COR translating at the same time [21], whereas the exo-knee joint only rotates around a fixed point. The inconsistency in knee joint motions is likely to cause sliding or small movement of tibia straps along the axial direction.

To model the directional differences in force responses, a tri-directional force element is introduced at each strap location. The force element measures three (XYZ) directional distances between a point on the exo-part and its counterpart on the body and generates positive or negative forces along these directions. In Figure 3, the two force elements located at the right femur and tibia strap locations are shown. In that figure, the green sphere illustrates a point on the exo-part and the purple wired sphere illustrates a nearby corresponding point on the human body. During the initial assembly, these two points are close to each other and generate zero force. The forces generated by a force element are modeled by linear damped springs:

$$\begin{cases} f_x = k_x(x - x_0) + c_x \dot{x} \\ f_y = k_y(y - y_0) + c_y \dot{y} \\ f_z = k_z(z - z_0) + c_z \dot{z} \end{cases} \quad (1)$$

The stiffness and damping constants of the four directional force elements are listed in Table 2. The stiffness in X direction is assumed to be 100 times of that in YZ direction to simulate the behavior of harder resistance in the fore-aft direction and softer resistance in the sliding and lateral directions. These parameters are chosen such that in the fore-aft direction (forward running direction) there should be little to no relative movement between strap and body, however relative motion is allowed in a sliding and lateral direction. And numerical tests with these parameters produced the desired outcome.

This tri-directional force element can be easily extended to a 6-direction force element such that two opposite directions along one axis can have different stiffnesses.

## 2.2 SIMULATION PARADIGM

Without a physical exoskeleton prototype, the experimental measurements of wearer's motions, GRFs and other interaction forces between the exoskeleton and the wearer are not available. Nonetheless, most data are readily available for normal motions (without exoskeleton) from motion capture experiments. Considering this, we assume the wearer is capable of maintaining the normal gait (with adapted effort) when wearing a properly designed exoskeleton, possibly after training. Therefore, the measured normal running motion is used to define the movement of the joints that are tracked within the simulations (with or without exoskeleton). This assumption is made as the objective of the work is to observe how the exoskeleton will impact GRF, joint torques, muscle activation, given that an optimized solution is obtained with the coordination between the actuators and human muscles that results in the normal motion. Unlike the joint kinematics, the GRFs measured from the normal gait are not directly applicable to exoskeleton simulations due to the weight difference. Therefore, the GRFs must be predicted in these simulations. Methods for GRF prediction from motion alone were proposed in the literature for walking and running [22, 23]. For the simulation of running, prediction of GRFs is very straightforward. Unlike a walking motion that has a double stance phase, running has only two phases: a single supporting phase and a flying phase. GRF prediction is only needed for the single supporting phase and it can be estimated through an equivalent force transformation method or an optimization method minimizing the difference between the GRF and the equivalent force with proper friction constraints [23]. The running motion data utilized in this study was collected in a study by Hamner et al. [24]. The subject with weight of 65.9kg ran at 3.96m/s (14.26km/h), three times his self-selected walking speed. With the assembled exoskeleton, the total weight is 88.9kg.

All simulations in this study were conducted with an in-house musculoskeletal simulation code, CoBi-Dyn, previously used for different studies [23, 25-27]. A hybrid inverse dynamics (ID) and forward dynamics (FD) simulation framework similar to the one presented in [25] was employed. The human body joints were classified as ID joints such that their motions can be prescribed to follow input experimental motion. The exoskeleton joints were classified as FD joints such that their motions were driven by the actuation forces and the wearer-exoskeleton interaction forces. At each time step, the hybrid dynamics framework predicted GRFs first and then joint torques for all human joints and accelerations for all exoskeleton joints. The predicted human joint torques were the target or desired torques that ideally shall be generated from muscles spanning these joints. In reality, the predicted joint torques could exceed the muscle strength or moment-generation capability. To compute muscle forces, the goal is to find an appropriate muscle force combination that generates the desired joint torques as closely as possible. Due to the redundancy in muscles, there could be numerous such combinations and thus muscle forces are determined by solving an optimization problem. The final objective of this optimization problem is to minimize an objective function, defined as $\sum_{i=1}^{n}\left(\frac{f_i}{f_i^{max}}\right)^p + w\boldsymbol{C}^T\boldsymbol{C}$, where $f_i$ is the force of the $i^{th}$ muscle, $f_i^{max}$ is the maximum attainable muscle force at its current state, $\boldsymbol{C}$ is the difference vector between the desired joint moments and the moments generated by spanning muscles (often called the residual torques); and $w$ is a weighting or penalty factor for the moment difference. $\frac{f_i}{f_i^{max}}$ can be considered as the muscle effort or activation equivalent. For all our simulations, $p = 2$ and $w = 100$ were utilized. The polynomial order $p$ is chosen based on the review of literature in [20, 28], and $w = 100$ is chosen because using a larger number does not seem to affect our simulation results. In general, the square of muscle activation is a relatively small number, and therefore it is not necessary to use a very large penalty factor for the moment difference term.

## 2.3 CONTROL SCHEME 1: ACTIVE TORQUE COMPENSATION CONTROLLER TO MINIMIZE INTERFERENCE (MIC)

First, we propose a simple torque compensation controller, called MIC, which aims to minimize the directional spring interaction force. Let's assume $\boldsymbol{\tau}_A$ and $\boldsymbol{\tau}_{SE}$ are the torques generated by the actuation forces and the spring forces (from the four springs on femurs and tibias) on the exoskeleton joints (exo-pelvis, exo-hip, and exo-knee), respectively. We have

$$\boldsymbol{\tau}_A = \boldsymbol{M}_A \boldsymbol{F}_A, \tag{2}$$

where $\boldsymbol{F}_A$ is the actuation force vector of dimension 6 and $\boldsymbol{M}_A$ is the generalized moment arm matrix (dimensions: $6 \times 6$) with respect to the exoskeleton joints. Since spring forces are passive and each is determined by its current length and velocity, $\boldsymbol{\tau}_{SE}$ can be computed as

$$\boldsymbol{\tau}_{SE} = \boldsymbol{M}_{SE}\boldsymbol{F}_S, \tag{3}$$

where $\boldsymbol{F}_S$ is the spring force vector of dimension 12 (4 springs with 3 variables each) and $\boldsymbol{M}_{SE}$ is the generalized moment arm matrix (dimensions: $6 \times 12$) with respect to the exoskeleton joints. Similarly, the torques generated from the spring forces with respect to human joints (3-DOF hip and 1-DOF knee, i.e., 4-DOF on each side) are

$$\boldsymbol{\tau}_{SH} = \boldsymbol{M}_{SH}\boldsymbol{F}_S, \tag{4}$$

where $\boldsymbol{M}_{SH}$ is a $8 \times 12$ moment arm matrix. Note $\boldsymbol{M}_H$, $\boldsymbol{M}_{SE}$ and $\boldsymbol{M}_{SH}$ all depend on the configuration or posture of the exoskeleton and the wearer, and they need to be updated when the configuration changes. With a physical exoskeleton prototype, variables such as $\boldsymbol{F}_S, \boldsymbol{F}_A, \boldsymbol{M}_A, \boldsymbol{M}_{SE}$ are readily measurable or can be easily computed from the exoskeleton sensory information alone. To determine $\boldsymbol{M}_{SH}$, the posture of the wearer at any time instance must be known. It can be estimated from additional sensors placed on the human body. It is also possible to utilize an inverse kinematics computation to estimate the wearer's posture from the XYZ distances of paired points in the four force elements (Figure 3), which are directly linked to $\boldsymbol{F}_S$. Therefore, all these quantities could be determined physically by the sensory information on the exoskeleton alone.

The goal of this first controller is to find the optimal actuation forces $\boldsymbol{F}_A$ (within limits) to minimize an objective function

$$\phi_1 = \boldsymbol{C}_1^T \boldsymbol{C}_1. \tag{5}$$

Here $C_1 = \tau_A - \tau_{SE}$ is the difference vector between the actuation and spring force torques. The idea is to produce an actuation torque that compensates the spring force torque and propels the exoskeleton to follow the wearer's motion closely with minimal spring forces. Minimizing the difference will in general produce $\tau_A$ in the same direction of $\tau_{SE}$, which means if $\tau_{SE}$ drags the exoskeleton forward so does $\tau_A$. In the pure passive mode, $\tau_A$ is a zero vector and the motion of the exoskeleton is mainly determined by $\tau_{SE}$ along with gravity and constraints. Minimizing $\phi_1$ leads to a $\tau_A$ that is close to $\tau_{SE}$ and it assists the exoskeleton motion with active actuation torques and reduces the spring forces. Note the spring forces are unlikely to vanish, which will cause $\tau_{SE} = 0$ and the optimization to predict zero assistance torque ($\tau_A = 0$). Therefore, this controller can only reduce interaction forces but not eliminate them.

## 2.4 CONTROL SCHEME 2: ACTIVE TORQUE COMPENSATION CONTROLLER TO MAXIMIZE ASSISTANCE (MAC)

The second controller, called MAC, aims to actively assist human motion via the spring interaction forces. At any instant, the required human joint torques at lower extremities, to be generated by muscle forces, are computed to track the target motion. This torque vector, $\tau_M$, is affected by the GRFs and spring forces. The goal is to reduce $\tau_M$ such that the muscle effort will be reduced significantly. One way to reduce $\tau_M$ is to make spring forces contribute positively to assist the motion. Let $\tau'_{SH} = M_{SH} F'_S$ be the torques generated from desired spring forces $F'_S$ on the human joints. Ideally, one would like to have $\tau_M = \tau'_{SH}$ such that the muscle contribution will not be needed. Without setting limits for these additional forces, $F'_S$ can be efficiently solved with the least square solution of this underdetermined system of equations. Next, we describe how to compute the actuation forces ($F_A$) to produce the desired $F'_S$.

Considering the torques generated by $F'_S$ on the exoskeleton joints, we have

$$\tau'_{SE} = M_{SE} F'_S. \tag{6}$$

This goal is to have the actuation torque $\tau_A$ to balance or compensate $\tau'_{SE}$ in the opposite directions such that if the spring forces drag exoskeleton backward the actuation forces will pull it forward. Therefore, we can define an objective function

$$\phi_2 = C_2^T C_2 \tag{7}$$

with $C_2 = \tau'_{SE} + \tau_A$. Minimizing $\phi_2$ will indirectly reduce $\tau_M$ and predict an optimal $F_A$ to assist the wearer's motion.

Both torque compensation controllers are used and based on minimizing an objective function for the torque differences between the actuator generated torques and the desired target torques (computed differently for the two controllers). Once the optimization is solved to obtain the actuator forces, the human joint torques required to track the normal running motion are computed. Given results of the human joint torques, muscle activations are calculated from the optimization presented earlier.

## 3. RESULTS

Four simulations were conducted and the results are presented here. We first conducted a simulation of the subject running normally (without the exoskeleton) to establish the baseline values of the biomechanical loadings (case 1). Then we conducted a simulation of the subject running with the exoskeleton in the fully passive mode (case 2), followed by two more simulations of the exoskeleton actively controlled with the controller 1 and 2 (case 3 and 4 respectively). Similar to the study by Uchida et al. [20], we assume the kinematics would change minimally during unassisted and assisted running and thus track the same running motion for all four cases. All results presented below are normalized by the gait cycle that started with the left foot impact and ended with the left foot impact again.

In Figure 4, snapshots of the muscle activation and the predicted GRFs from the simulation of a full running gait cycle with exoskeleton (control 2) are shown. In Figure 5, the predicted GRFs are shown for all four simulations. Apparently, the vertical GRF is much higher than other components. For the normal running case, the predicted vertical GRF is very close to the measured value with a peak force around 1536N. A detailed comparison of the predicted GRFs with the experimental measured GRFs for the normal running case can be found in [23]. The vertical GRFs for the three exoskeleton simulations are relatively close and the peak values are 2015, 2018, 2042N, for passive, control 1 and control 2 cases, respectively. The peak value of the passive exoskeleton ($2015N$) is 31% more than that of the normal running, close to the percent increase of weight due to exoskeleton ($23/65.9 \cong 35\%$). Like the vertical GRFs, the fore-aft GRFs all share similar patterns and the peak values for exoskeleton simulations are close, an increase of around 34% from the normal running simulation. Nonetheless, for the lateral GRFs, the pattern for control 2 is quite different from others, with its peak value of around 117N compared to less than 40N for others. In addition, the peak force happens mostly during the second half of the single stance phase.

In Figure 6, the joint torques for hip and knee are plotted. During the single stance phase, the knee requires mostly extension torque (to straighten the knee or swing forward) and the hip requires extension first (backward swing) and then flexion (forward swing). The peak hip flexion, extension, abduction, and rotation torques and the peak knee extension torque are summarized in Table 3 for all four cases. When compared to the normal running case, the passive exoskeleton increases the knee torque by 43% and the hip flexion, extension, abduction, and rotation torques by 39%, 32%, 61%, and 40%, respectively. The control 1 reduces all hip joint torques by less than 15% when compared to the passive exoskeleton but increases the knee joint torque slightly (1%). When compared to the normal running, it increases the hip joint torques by 37% at its highest (for abduction) and the knee extension torque by 44%. In contrast, the control 2 reduces the knee extension torque by 28%, hip flexion by 31%, and hip extension by 38%, when compared to the normal running. It somehow increases the abduction torque by $1 Nm$ (1%) and the rotation torque by $10 Nm$ (21%), which could be explained by the lack of hip rotation DOF for the exoskeleton. When compared to the passive exoskeleton case, the control 2 reduces the knee extension torque by almost a half (49%) and reduces the hip joint torques across the board, range from 13% (hip rotation) to 53% (hip extension).

In Figure 7 and Figure 8, the femur and tibia spring forces along all three directions are plotted. The spring forces for control 1 largely follow the same pattern as the passive exoskeleton but with much smaller magnitude, indicating less interference from the exoskeleton to the wearer. The forces along the vertical and lateral directions are much smaller than those in the fore-aft direction due to the stiffness difference. In the wide range of the gait cycle, the spring forces from control 2 have opposite signs from the passive exoskeleton or control 1, which clearly indicates active assistance instead of interference is fed to human motion. In the fore-aft direction, a large positive spring force (838N) is predicted for the control 2 simulation.

In Figure 9, the optimized actuation forces for the three actuators (exo-pelvis, exo-hip, exo-knee) are shown for both controllers. The control 2 actuation forces have similar pattern as the control 1 forces but with much bigger magnitude. The negative actuation force around the knee means it pushes to extend the knee. The hip actuation force is negative first (push to extend the hip) and then positive (pull to flex hip) during the stance phase. The knee and hip forces follow a similar pattern as the torques generated by muscles in Figure 6. The pelvis actuation force is mostly positive which means it pulls to rotate the exo-pelvis part to help abduct the hip and therefore its pattern is likely to follow that of hip abduction torque in Figure 6 (with sign difference).

Based on the predicted joint torques in Figure 6 and subsequent muscle force optimization, muscle forces and activations were obtained. In Figure 10, muscle activations of six selected muscles (on the left leg) spanning the hip and kneed joints are compared in all four cases. The predicted muscle activations for the normal running are similar in trend as the experimental EMG presented in [24, 29] or the predicted muscle activations in [20, 24]. Compared to the normal running, muscle activations increase considerably for the passive exoskeleton case and the exoskeleton case with the controller 1. The largest increase is observed for the rectus femoris muscle with its activation almost doubled. The vastus lateralis muscle has the smallest change in activation because its activations almost reach the maximum in all cases, including the normal running case. For the biceps femoris long and short heads and gluteus maximum muscles, the activations increase by less than 30% mostly. Compared to the passive exoskeleton case, the case with active controller 1 reduces muscle activations to a modest degree for all muscles except vastus lateralis and medialis. Comparing to the normal running case, the control 2 reduces muscle activations in all six muscles. The reduction for vastus lateralis is minor since the peak is almost the same despite the pulse width being slightly narrower. For the other five muscles, the activation reductions are evident. Among them, the biceps short head muscle sees the largest reduction in peak activation from 0.97 to 0.3. Another two muscles, gluteus maximus and rectus femoris, also see their peak activations reduced by more than a half.

Due to changes in muscle contractions, the joint reaction forces vary accordingly. Figure 11 shows the comparison of knee joint reaction forces along the tibia axes. During normal running, the maximal axial force on knee is around $4605N$. In [30], the knee loading during jogging is measured to be around $3000N$ when normalized to average body weight of $75kg$, based on instrumented knee implant measurements of 3 subjects jogging at $6km/h$. The highest knee loading measured during slow jogging is up to $5,165N$. Our predicted force is higher than the average value. The discrepancy is likely due to substantial differences between the participants and their gait characteristics. The data for current study was collected from a young adult while the direct measurements were from elderly adults with total knee replacement. In addition, the running speed for our subject is much higher at $14.26$ $vs$ $6 km/h$. The predicted joint forces for passive exoskeleton and control 1 are both significantly higher than those of the normal running with peak forces around $7200N$. The control 2 reduces the peak force to around $6340N$ but is still greater than the normal running, which can be attributed to the added weight from the exoskeleton despite the torque assistance it provides.

## 4. DISCUSSION AND CONCLUSIONS

The exoskeleton presented in this study is relatively heavy (23 kg) but within a reasonable range. For example, the exoskeleton prototype evaluated by Gregorczyk et al [3] weights 15kg and can carry extra heavy backpack loading. The HAL system [1] weighs 23kg and includes an on-board battery. Other systems such as the SARCOS XOS are much bulkier. More recent electrically actuated lower limb exoskeletons are likely to be lighter. This weight may affect the operation of the exoskeleton and the requirement of the maximum actuation forces. For the current design, the 4000N actuation force limit set for all actuators was shown to be strong enough to provide desired assistive torques for all assisted joints (Figure 9). Nonetheless, it also generated relatively large human-device interaction forces, especially along the fore-aft direction. As shown in Figure 7 and Figure 8, the maximum fore-aft forces acted on the upper and lower legs are over 800N and 500N, respectively. Assuming a strap contact area of $200 cm^2$, the interaction force can cause skin pressure over $40 kPa$. This pressure is much smaller than the instant pressure pain threshold at around 280 to 480kPa [31, 32]. In the work by Tamez-Duque et al. [31], the maximum average of thigh strap pressure measured during walking with a powered exoskeleton reach $37 kPa$ for a spinal cord injury patient. At this pressure level, prolonged use of the powered exoskeleton can still cause skin discomfort and may restrict blood flow that can potentially lead to skin ulcers and infections. The strap-skin pressure during running is expected to be higher than that of walking. It can be potentially lowered by reducing the actuation assistance, adjusting strap tension or stiffness, or increasing the contact area between the leg and strap. In addition, it is possible to reduce the strap forces on the legs by loosening the tie constraint between the load support and the human body to allow small relative movement. However, doing so will increase the complexity of modeling and require additional numerical tests to calibrate interaction force parameters.

The actuation models, represented as yellow cylinders in Figure 1, are idealized and simplified. Accessories associated with the actuators, such as motors, transmissions, wires, and batteries, are not represented but their masses are lumped into the exo-parts. Some other critical design issues may have been ignored as well. For example, if the actuators are hydraulic, the dynamics of the actuators and linkages should be considered. However, it may be difficult to install hydraulic cylinders so close to human body, whereas electric motor actuators could fit better.

In the study by Uchida et al. [20], ideal and massless assistive devices were added to hips, knees, and ankles to generate muscle-driven simulations of running. They found these ideal joint torque assistances in general are effective in reducing the activations and metabolic powers of muscles crossing assisted joints. Like their study, our work assumes the kinematics would change minimally during unassisted and assisted running (due to the lack of test data for running with the exoskeleton). On the other hand, we modeled a multi-joint exoskeleton explicitly in this study and estimated the force interactions between the wearer and the exoskeleton. The GRFs predicted are affected by the added exoskeleton mass and assistance, unlike the ideal assistance in [20]. And the detrimental effects of a heavy (passive) exoskeleton on the wearer's musculoskeletal loadings are clearly shown. For the two controller cases, for which the 6 force actuators are used to provide indirect assistance to joints, we observed the effects on reducing muscle activations. However, we also discovered that the axial joint reaction force increases due to added exoskeleton weight, which is unlikely to be true for massless devices. This implies that, when evaluating a wearable exoskeleton, we also need to pay attention to its effects on joint reaction forces and potential negative impacts on injury risks.

Both controllers presented in this paper rely on the human-device interaction forces as sensory information for the prediction of actuation forces. The interaction forces can be potentially measured with tri-axial load cells [33]. However, noise and drifting from the sensor measurement can compromise the performance of the controller. In practice, additional steps to filter and recalibrate the force measurement online should be considered to improve the robustness of these controllers. Besides the interaction forces, the first controller only needs positional information of the exoskeleton (i.e. exo-joint angles) to function, which can be measured with joint encoders or other devices. For the second controller, estimation of the lower extremity joint torques is needed that often requires the knowledge of human joint kinematics and GRFs. Consequently, practical implementation of this controller will be more complicated than the first one. Wearable inertial measurement units (IMUs) can often be used to estimate human poses or motions. For example, von Marcard et al. [34] used as few as 6 IMUs to estimate human poses for arbitrary human motions. Together with GRF measurement, the joint torques can be estimated through inverse dynamics. In a recent study [35], it was shown ankle torque can be estimated without the need for horizontal GRFs. Considering this, GRF measurement with force insoles [36], which typically can only measure vertical GRFs, could be a feasible option. On the other hand, due to recent progresses in data driven machine learning research, it might be possible to estimate joint torques directly from IMU measurements [37]. And the accuracy of such estimation likely depends on the amount of ground truth (training) data available.

The current study utilizes only a running motion to analyze the human exoskeleton interaction. During running, the GRFs need to be predicted only for the single stance phase, which can be computed efficiently and accurately [23]. In contrast, for a walking gait, there is a double stance phase for which the prediction of GRFs becomes an indeterminacy problem and requires more complex algorithms such as optimization and assumption of smooth transition [38]. Incorporating such methods for predicting GRFs during the double stance phase in our simulation framework can enable analyses of human exoskeleton interactions during walking.

The current exoskeleton design represents a body-worn device that has no direct contact with the ground and provides no assistance to the ankle. Considering the important role of the ankle during gaits, a potential future work is to evaluate an exoskeleton design with an exo-foot component and linked actuators to assist each ankle. The exo-foot components instead of human feet can come to direct contact with the ground and can provide the mechanism to transfer the exoskeleton weight or load to the ground without adding too much burden to the wearer. However, such a design likely will affect the wearer's gait more, which means using a normal running motion may not predict accurate results.

In this study, the simulations were conducted only for one individual with a properly fitted exoskeleton. To consider multiple subjects, it likely requires modification to the design or dimensions of the skeleton to ensure proper fit with different subjects for optimal performance. Therefore, the scope of this work is limited to demonstrate the feasibility of using the human-in-the-loop simulation paradigm for exoskeleton design and evaluation, although completing the analysis on multiple individuals would improve confidence in the findings.

In conclusion, we presented an integrated human-in-the-loop simulation paradigm for the design and evaluation of two virtual torque compensation controllers for a lower extremity exoskeleton. The two controllers, aiming to reduce interference and to provide assistance, are straightforward to implement numerically and can be potentially transferred to physical prototypes without difficulty. Our simulations have provided sound evidence of the efficacy of these controllers, by examining the exoskeleton-wearer interaction forces, human joint torques and joint reaction forces, and by comparing them with those of the passive exoskeleton. The second assistive controller in particular reduces both hip and knee joint torques substantially as shown by the results.

Nonetheless, the knee joint reaction force still increases when compared to normal running due to the added weight. The present simulation paradigm can be utilized to evaluate the design of exoskeletons and control schemes and to predict their effects on human biomechanical loadings. Parametric simulations can also be performed to optimize design parameters such as the strap tri-directional stiffness and exo-part dimensions. The present simulations and knowledge gained can benefit the development of extensions to the simulation method and provide guidelines for building novel exoskeleton prototypes.

**Figure Captions List**

Fig. 1 The lower extremity exoskeleton design with its joint axes shown. The yellow cylinders are the actuators. The right two figures show joint movement of the mechanism and the rightmost one shows the two exo-pelvis parts (green) and the actuators (yellow) that drive abduction/adduction.

Fig. 2 The lower extremity exoskeleton assembled onto the human body model.

Fig. 3 (a) The directional force elements at the right strap locations. (b) Zoom-in view of the tibia force element.

Fig. 4 Snapshots from the simulation of a running gait cycle with exoskeleton and control 2 method. The muscle color indicates its activation and purple arrows are predicted GRFs.

Fig. 5 Predicted ground reaction forces (GRFs).

Fig. 6 Comparison of hip and knee torques generated by muscles. Note the sign of the torques for hip flexion (+: flexion; -: extension); hip abduction (+: adduction; -: abduction), and knee extension (+: extension; -: flexion).

Fig. 7 Femur spring forces along three directions. The positive fore-aft force helps to extend the hip and positive lateral force helps to abduct the hip.

Fig. 8 Tibia spring forces along three directions. The negative fore-aft force helps to extend the knee and positive lateral force helps to abduct the hip.

Fig. 9  Active actuation forces. Positive force indicates pulling whereas negative force indicates pushing. The force ranges along the vertical axes are the same for all three plots.

Fig. 10  Comparison of muscle activations predicted from all four simulation for six major muscles around the hip and knee.

Fig. 11  Axial joint reaction forces at knee. Negative value indicates compression.

**Table Caption List**

Table 1   Mass and inertia properties of exoskeleton components. x: fore-aft; y: vertical; z: lateral.

Table 2   Stiffness and damping of the direction springs.

Table 3   Peak hip and knee joint torques (unit: Nm) predicted for all four cases. The first percent value in the brackets is relative to the normal running case (No Exo) and the second one is relative to the passive exoskeleton case (Passive).

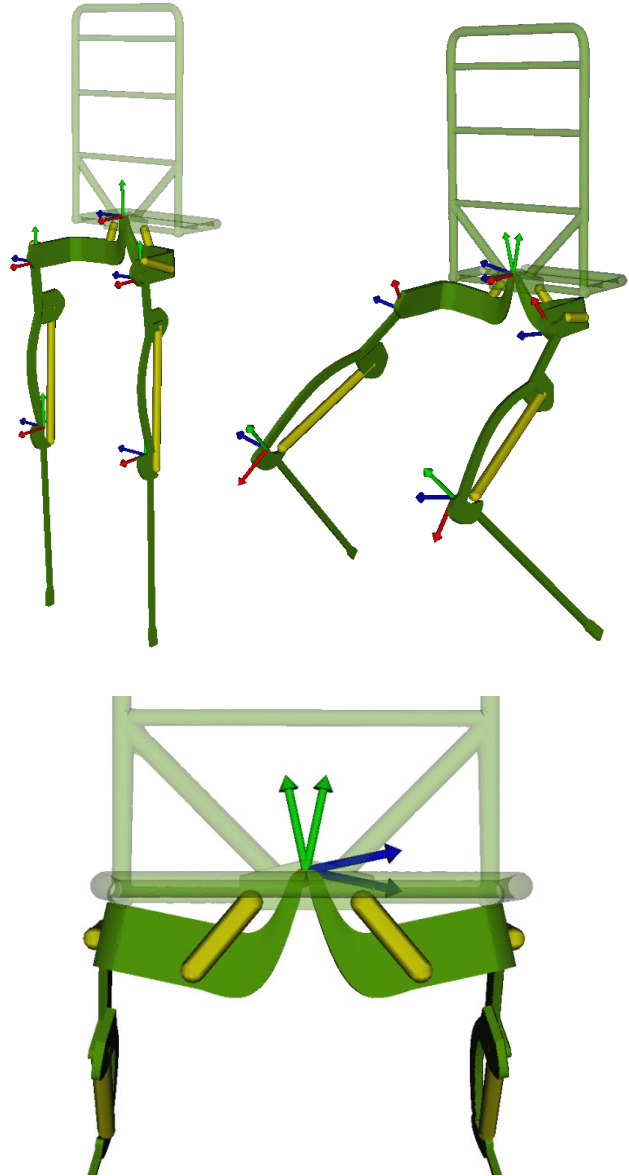

*Figure 1. The lower extremity exoskeleton design with its joint axes shown. The yellow cylinders are the actuators. The right two figures show joint movement of the mechanism and the rightmost one shows the two exo-pelvis parts (green) and the actuators (yellow) that drive abduction/adduction.*

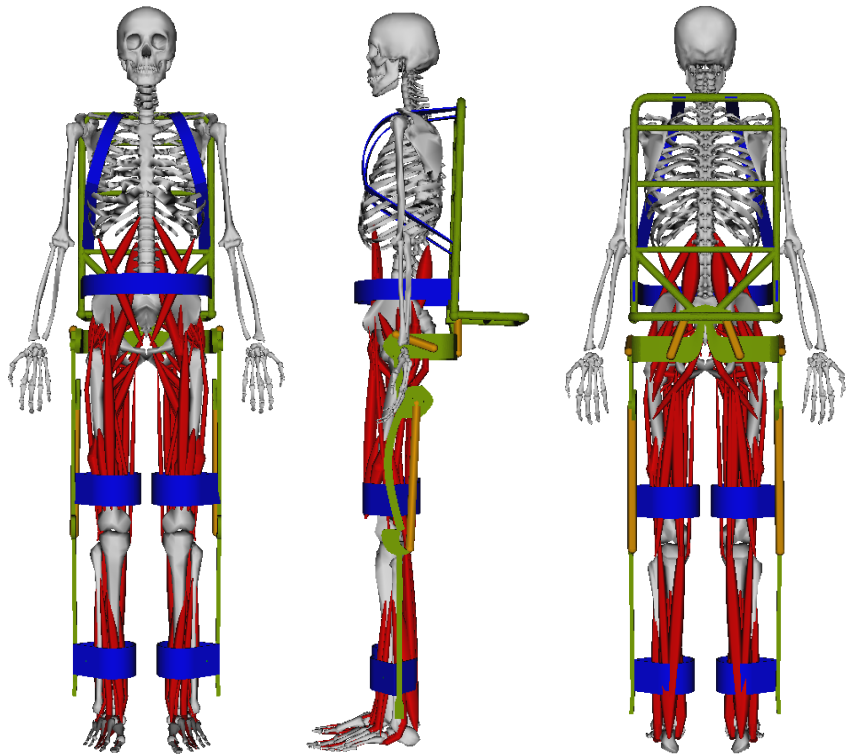

*Figure 2. The lower extremity exoskeleton assembled onto the human body model.*

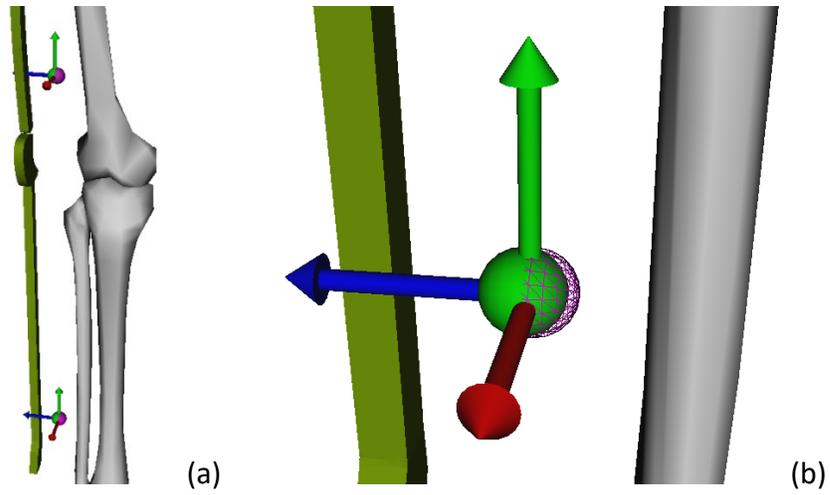

*Figure 3. (a) The directional force elements at the right strap locations. (b) Zoom-in view of the tibia force element.*

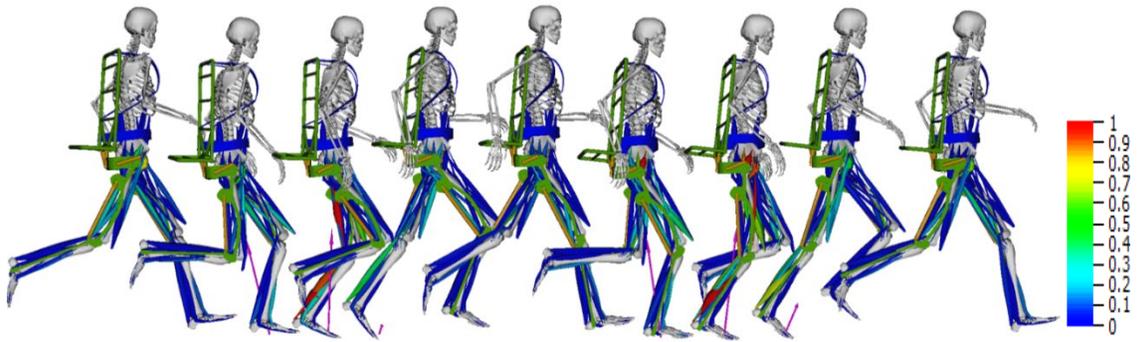

*Figure 4. Snapshots from the simulation of a running gait cycle with exoskeleton (control 2) method. The muscle color indicates its activation and purple arrows are predicted GRFs.*

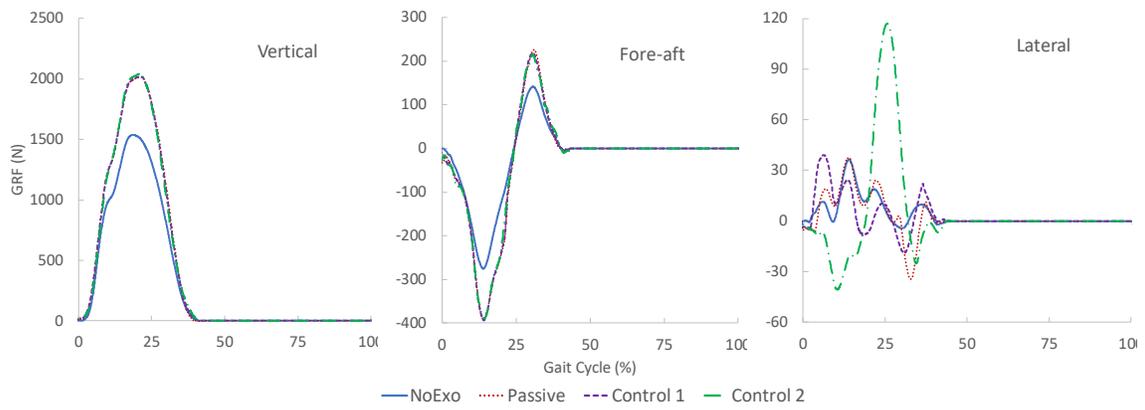

*Figure 5. Predicted ground reaction forces (GRFs).*

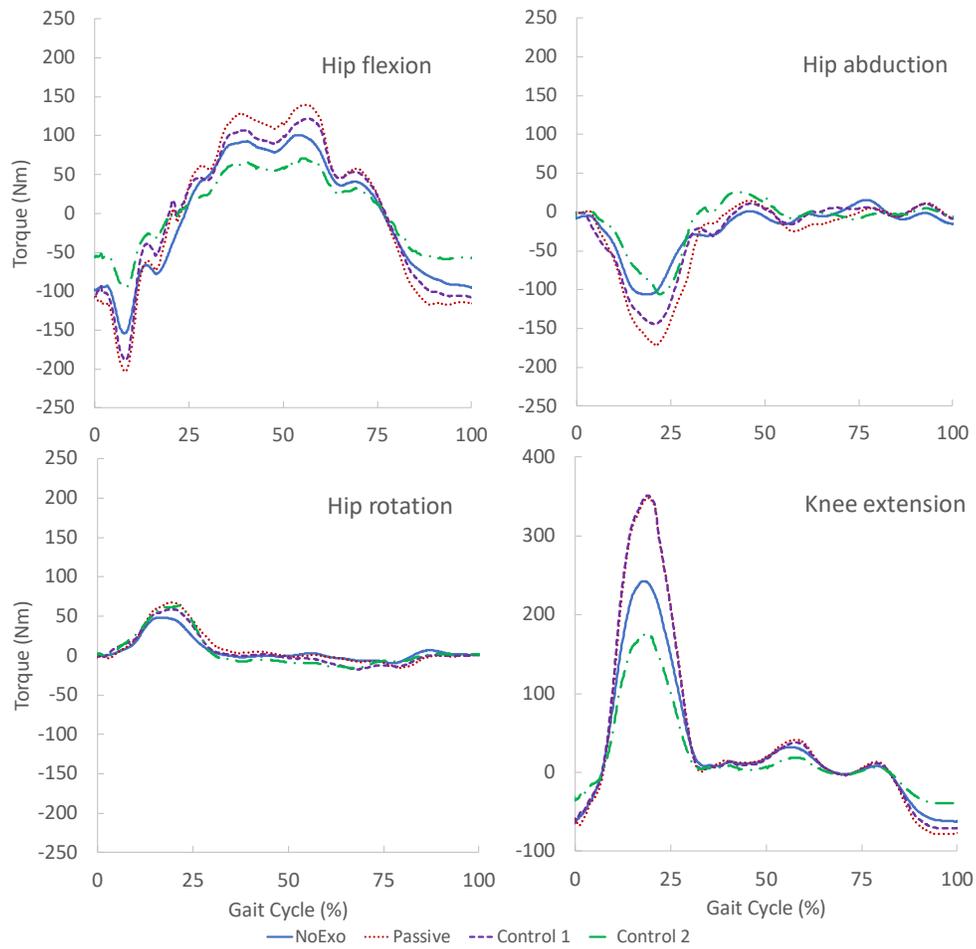

*Figure 6. Comparison of hip and knee torques. Note the sign of the torques for hip flexion (+: flexion; -: extension); hip abduction (+: adduction; -: abduction), and knee extension (+: extension; -: flexion).*

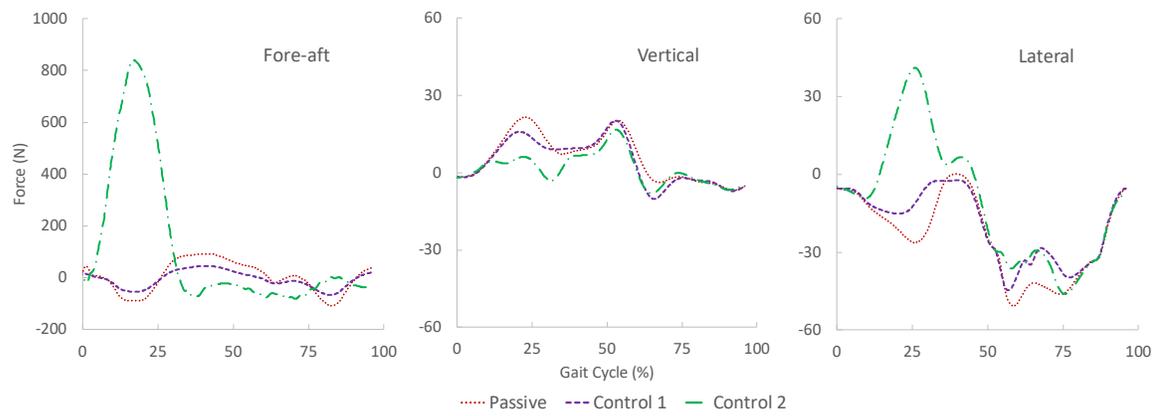

*Figure 7. Femur spring forces along three directions. The positive fore-aft force helps to extend the hip and positive lateral force helps to abduct the hip.*

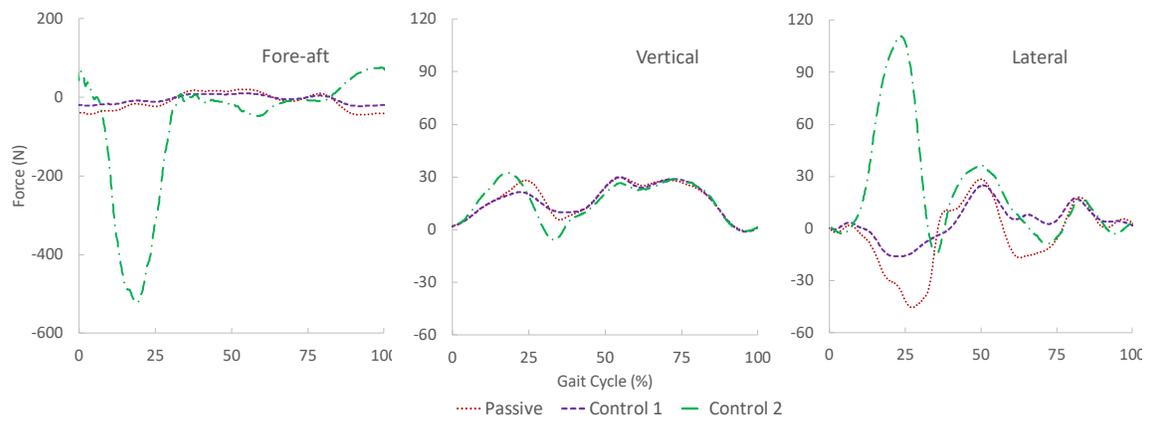

*Figure 8. Tibia spring forces along three directions. The negative fore-aft force helps to extend the knee and positive lateral force helps to abduct the hip.*

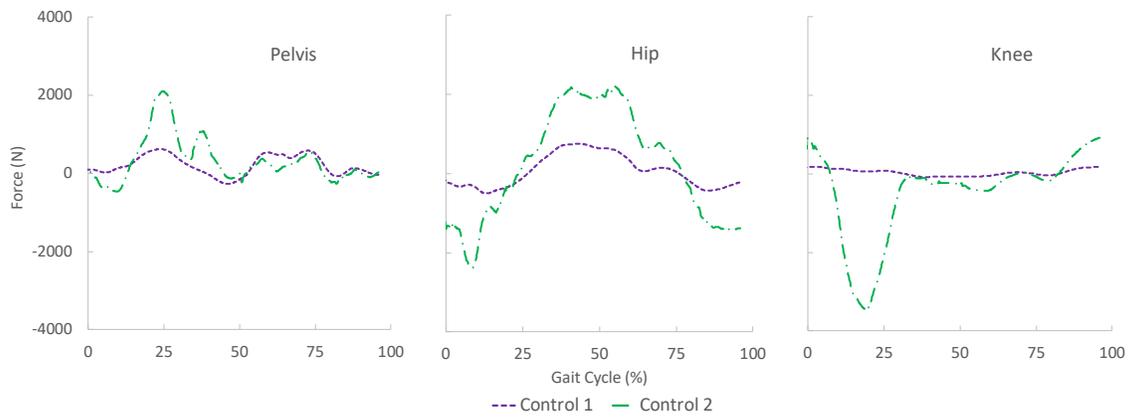

*Figure 9. Active actuation forces. Positive force indicates pulling whereas negative force indicates pushing. The force ranges along the vertical axes are the same for all three plots.*

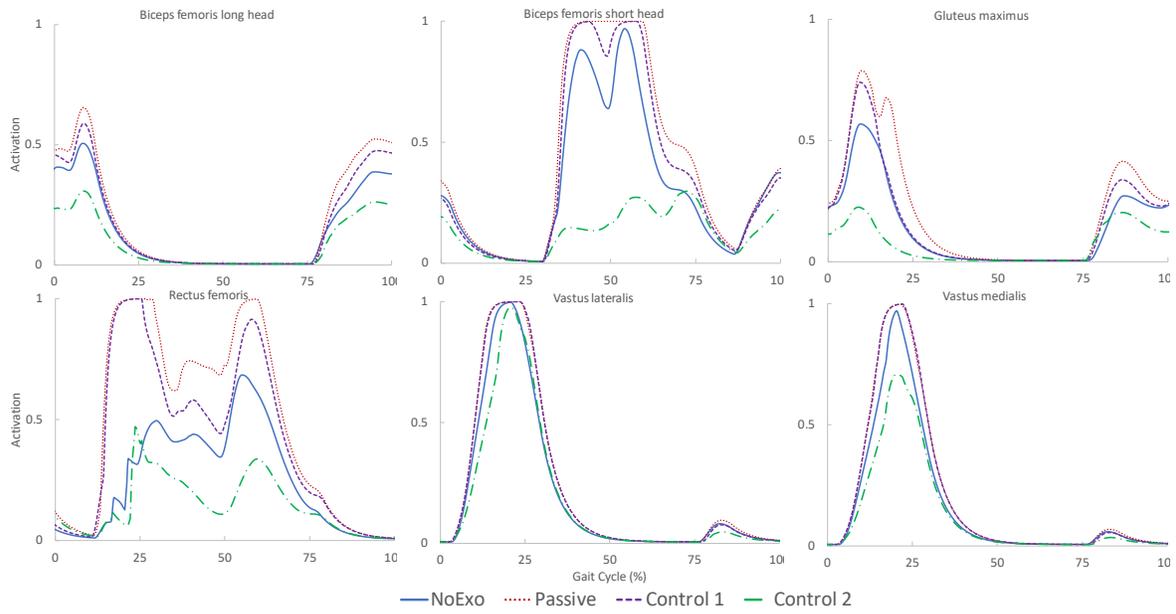

*Figure 10. Comparison of muscle activations predicted from all four simulation for six major muscles around the hip and knee.*

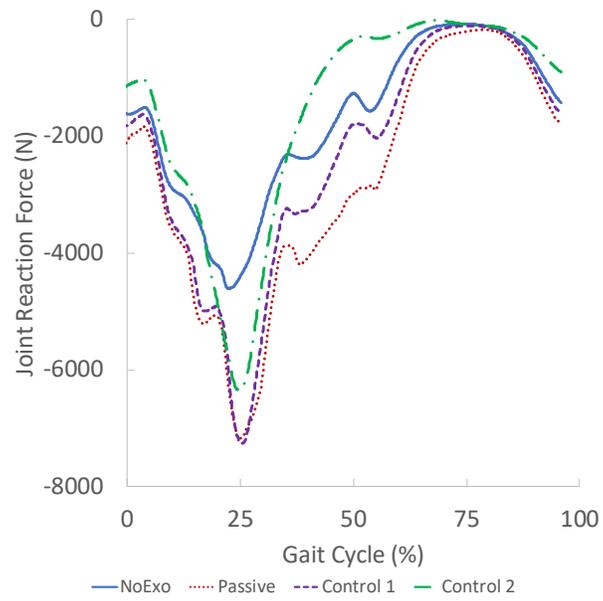

*Figure 11. Axial joint reaction forces at knee. Negative value indicates compression.*

*Table 1. Mass and inertia properties of exoskeleton components. x: fore-aft; y: vertical; z: lateral.*

| Exo-Component | Mass (kg) | $I_{xx}$ | $I_{yy}$ | $I_{zz}$ |
|---|---|---|---|---|
| Load Support | 3 | 0.150 | 0.050 | 0.110 |
| Pelvis (L/R) | 5 | 0.0181 | 0.0311 | 0.0172 |
| Femur (L/R) | 3 | 0.0640 | 0.0011 | 0.0640 |
| Tibia (L/R) | 2 | 0.0420 | 0.0007 | 0.0420 |

*Table 2. Stiffness and damping of the direction springs.*

|  | Stiffness (N/m) | | | Damping (Ns/m) | | |
| --- | --- | --- | --- | --- | --- | --- |
|  | $k_x$ | $k_y$ | $k_z$ | $c_x$ | $c_y$ | $c_z$ |
| Femur | 160000 | 1600 | 1600 | 400 | 40 | 40 |
| Tibia | 160000 | 1600 | 1600 | 400 | 40 | 40 |

*Table 3. Peak hip and knee joint torques (unit: Nm) predicted for all four cases. The first percent value in the brackets is relative to the normal running case (No Exo) and the second one is relative to the passive exoskeleton case (Passive).*

|  | Hip | | | | Knee |
| --- | --- | --- | --- | --- | --- |
|  | Flexion | Extension | Abduction | Rotation | Extension |
| No Exo | 101 | 154 | 106 | 48 | 243 |
| Passive | 140 | 203 | 171 | 67 | 347 |
|  | (39%, —) | (32%, —) | (61%, —) | (40%, —) | (43%, —) |
| Control 1 | 122 | 188 | 145 | 63 | 350 |
|  | (21%, -13%) | (22%, -7%) | (37%, -15%) | (31%, -6%) | (44%, 1%) |
| Control 2 | 70 | 95 | 107 | 58 | 176 |
|  | (-31%, -50%) | (-38%, -53%) | (1%, -37%) | (21%, -13%) | (-28%, -49%) |